\pdfoutput=1

\documentclass[11pt]{article}

\usepackage{EMNLP2022}

\usepackage{times}
\usepackage{latexsym}

\usepackage[T1]{fontenc}

\usepackage[utf8]{inputenc}

\usepackage{microtype}

\usepackage{inconsolata}

\usepackage{graphicx}
\usepackage{algorithm}
\usepackage{algorithmic}

\usepackage{amsmath}
\usepackage{amssymb}

\usepackage{multirow}
\usepackage{float}

\usepackage{subfig}
\usepackage{makecell}

\newcommand \footnoteONLYtext[1]
{
	\let \mybackup \thefootnote
	\let \thefootnote \relax
	\footnotetext{#1}
	\let \thefootnote \mybackup
	\let \mybackup \imareallyundefinedcommand
}
\title{A Span-level Bidirectional Network for Aspect Sentiment Triplet Extraction}

\author{Yuqi Chen, Keming Chen \thanks{*Corresponding author}, Xian Sun\and Zequn Zhang \\
	Aerospace Information Research Institute \\ 
	Key Laboratory of Network Information System Technology(NIST) \\ 
	School of Electronic, Electrical and Communication Engineering\\ University of Chinese Academy of Sciences \\ 
	\texttt{chenyuqi19@mails.ucas.ac.cn}, \texttt{ckmdejob@hotmail.com}, \{sunxian,zqzhang1\}@mail.ie.ac.cn}

\begin{document}
\maketitle

\begin{abstract}
Aspect Sentiment Triplet Extraction (ASTE) is a new fine-grained sentiment analysis task that aims to extract triplets of aspect terms, sentiments, and opinion terms from review sentences. Recently, span-level models achieve gratifying results on ASTE task by taking advantage of the predictions of all possible spans.  Since all possible spans significantly increases the number of potential aspect and opinion candidates, it is crucial and challenging to efficiently extract  the triplet elements among them. In this paper, we present a span-level bidirectional network which utilizes all possible spans as input and extracts triplets from spans bidirectionally. Specifically, we devise both the aspect decoder and opinion decoder to decode the span representations and extract triples from aspect-to-opinion and opinion-to-aspect directions. With these two decoders complementing with each other, the whole network can extract triplets from spans more comprehensively. Moreover, considering that mutual exclusion cannot be guaranteed between the spans, we design a similar span separation loss to facilitate the downstream task of distinguishing the correct span by expanding the KL divergence of similar spans during the training process; in the inference process, we adopt an inference strategy to remove conflicting triplets from the results base on their confidence scores. Experimental results show that our framework not only significantly outperforms state-of-the-art methods, but achieves better performance in predicting triplets with multi-token entities and extracting triplets in sentences contain multi-triplets\footnotemark. 
\end{abstract}

$\footnoteONLYtext{This paper is accepted as a long paper in EMNLP 2022.}$
$\footnotetext{We release our code at \url{https://github.com/chen1310054465/SBN}}$

\section{Introduction}

\begin{figure} [t]
	\centering
	\includegraphics[scale=1.3]{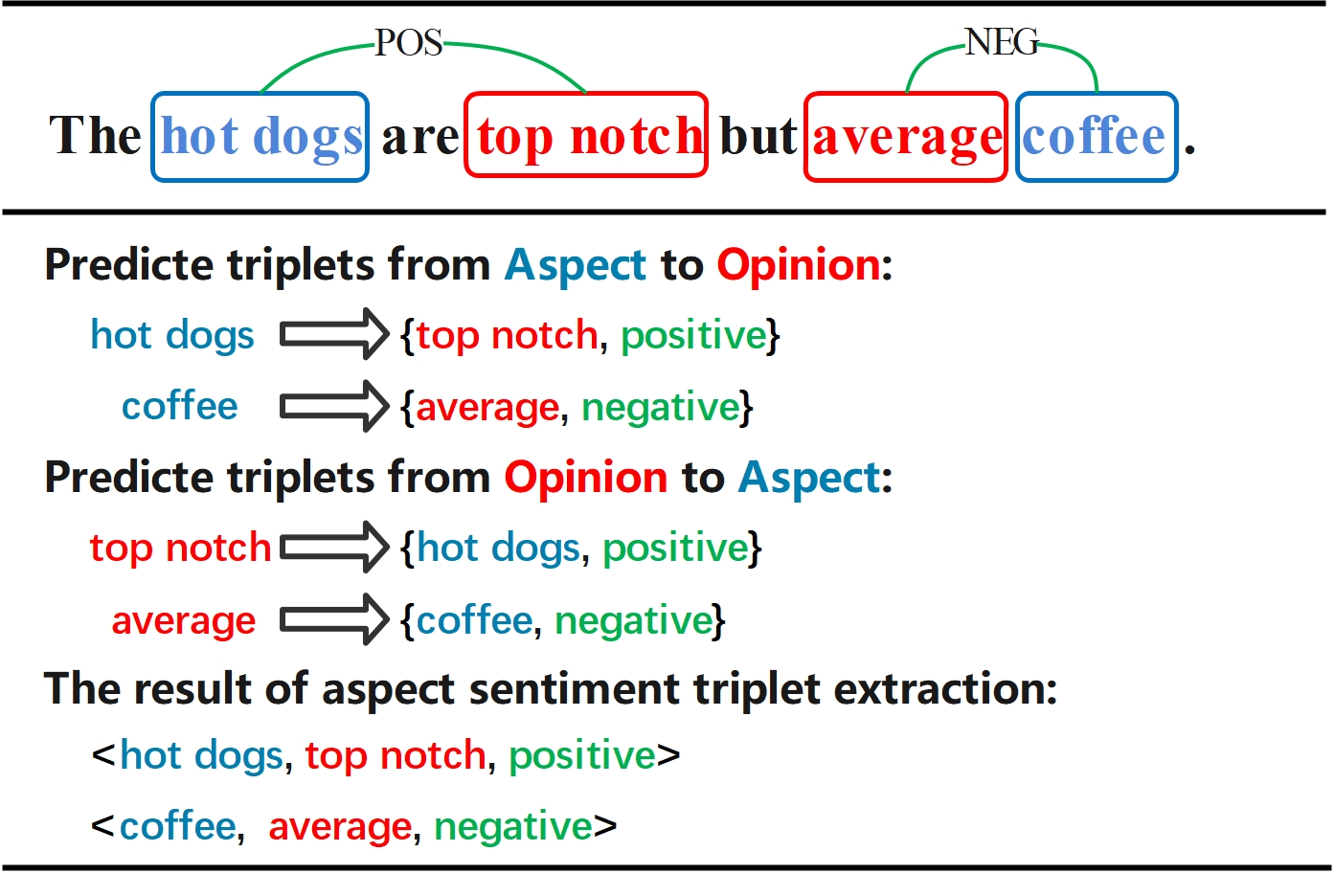}
	\caption{An example of ABSA subtasks. The spans highlighted in blue are aspect terms. The spans in red are opinion terms. Sentiments are marked with green.}
	\label{example}
\end{figure}

Aspect-based sentiment analysis (ABSA) is an important field in natural language processing (NLP). The ABSA task contains various fundamental subtasks, such as aspect term extraction (ATE), opinion term extraction (OTE), and aspect-level sentiment classification (ASC). Recent studies focus on solving these tasks individually or doing a combination of two subtasks, such as aspect term polarity co-extraction (APCE), aspect opinion co-extraction (AOCE), and aspect-opinion pair extraction (AOPE). However, none of these subtasks aims to extract the aspect terms (AT) with their corresponding opinion terms (OT) and sentiment polarity (SP) simultaneously. To tackle this problem, \cite{DBLP:conf/aaai/PengXBHLS20} propose the aspect sentiment triplet extraction (ASTE) task which aims to extract (\emph{AT, OT, SP}) triplets such as (\emph{hot dogs, top notch, positive}) and (\emph{coffee, average, negative}) in the example of Figure \ref{example}. 

To solve the ASTE task, recent works \cite{DBLP:conf/aaai/PengXBHLS20,DBLP:journals/corr/abs-2010-04640,DBLP:conf/aaai/MaoSYC21} use sequential token-level methods and formulate this task as a sequence tagging problem. Although these works achieve competitive results, their token-level models suffer from cascading errors due to sequential decoding. Therefore, \cite{DBLP:conf/acl/XuCB20} propose a span-level model to capture the span-to-span interactions among ATs and OTs by enumerating all possible spans as input. Despite the exciting results their work has yielded, several challenges remain with the existing span-level model. \textbf{First}, since both aspect terms and opinion terms can trigger triplets, it is a challenge to identify triplets bidirectionally. \textbf{Second}, unlike token-level methods, span-level input cannot guarantee mutual exclusivity among the spans, so the similar spans (spans that have shared tokens) such as \emph{hot dogs}, \emph{dogs}, and \emph{the hot dogs}, may cause confusion in downstream tasks. Thus, it is challenging for span-level models to effectively distinguish these similar span. \textbf{Third}, the existence of similar spans enables span-level models to generate conflicting triples in the results, such as (\emph{hot dogs, top notch, positive}), (\emph{hot dogs,are top notch, positive}), and (\emph{the hot dogs, top notch, positive}). How to properly extract non-conflicting triplets is also challenging.

To address these challenges, we propose a span-level bidirectional network for ASTE task. Unlike prior span-level works \cite{DBLP:conf/acl/XuCB20}, our network decodes all possible span representations from both aspect-to-opinion and opinion-to-aspect directions through the cooperation of the aspect decoder and opinion decoder. In the aspect-to-opinion direction, the aspect decoder aims to extract ATs such as $\left\{\emph{hot dogs, coffee}\right\}$, and the opinion decoder aims to extract OTs such as $\left\{\emph{top notch}\right\}$ for each specific AT like $\left\{\emph{hot dogs}\right\}$. Analogously, in the opinion-to-aspect direction, the opinion decoder and aspect decoder are utilized to extract OTs and their corresponding ATs, respectively. Furthermore, we design the similar span separation loss to direct the model deliberately distinguishing similar span representations during the training process; and an inference strategy employed in the prediction process is also proposed for eliminating the conflicting triplets in the extraction results. To verify the effectiveness of our framework, we conduct a series of experiments based on four benchmark datasets. The experimental results show our framework substantially outperforms the existing methods. In summary, our contributions are as follows:
\begin{itemize}
	
	\item We design a span-level bidirectional network to extract triplets in both aspect-to-opinion and opinion-to-aspect directions in a span-level model. By this design, our network can identify triplets more comprehensively.
	
	\item We propose the similar span separation loss to separate the representations of spans that contain shared tokens. Based on these differentiated span representations, downstream models can discriminate the span representation more precisely.
	
	\item We design an inference strategy to eliminate the potential conflicting triplets due to the lack of mutual exclusivity among spans.
\end{itemize}

\section{Related Work}

Aspect based sentiment analysis (ABSA) is a fine-grained sentiment analysis task that consists of various subtasks, including aspect term extraction (ATE) \cite{DBLP:conf/emnlp/WangPDX16, DBLP:conf/emnlp/LiL17, DBLP:conf/acl/XuLSY18, DBLP:conf/ijcai/LiBLLY18, DBLP:conf/acl/MaLWXW19}, opinion term extraction (OTE) \cite{DBLP:journals/kbs/PoriaCG16, DBLP:conf/naacl/FanWDHC19,DBLP:journals/corr/abs-2010-04640}, aspect-level sentiment classification (ASC) \cite{DBLP:conf/acl/DongWTTZX14, DBLP:conf/emnlp/TangQL16, DBLP:conf/acl/HeLND18, DBLP:conf/aaai/LiW0Z019}. Since these subtasks are solved individually, recent studies attempted to couple two subtasks as a compound task, such as aspect term polarity co-extraction (APCE) \cite{DBLP:conf/aaai/LiL17, DBLP:conf/acl/HeLND19,  DBLP:conf/aaai/LiBLL19}, aspect and opinion co-extraction \cite{DBLP:journals/coling/QiuLBC11, DBLP:conf/ijcai/LiuXLZ13, DBLP:journals/taslp/YuJX19}, aspect category and sentiment classification \cite{DBLP:conf/emnlp/HuZZCSCS19}, and aspect-opinion pair extraction (AOPE) \cite{DBLP:conf/acl/ChenLWZC20, DBLP:conf/acl/ZhaoHZLX20, DBLP:conf/aaai/GaoWLWZL21}, and aspect-opinion pair extraction (AOPE) \cite{DBLP:conf/aaai/GaoWLWZL21, DBLP:conf/acl/ZhaoHZLX20, DBLP:conf/ijcai/Wu0RJL21}. Although many works have achieved great progress on these tasks, none of these tasks aims to identify the aspect terms as well as their corresponding opinion term and sentiment polarity.

To tackle this issue, \cite{DBLP:conf/aaai/PengXBHLS20} proposed the aspect sentiment triplet extraction (ASTE) task, which aimed to extract aspect terms, the sentiments of the aspect terms, and the opinion terms causing the sentiments. Some methods \cite{DBLP:conf/emnlp/XuLLB20,DBLP:journals/corr/abs-2010-04640} designed a unified tagging scheme to solve this task. Some others \cite{chen2021bidirectional, DBLP:conf/aaai/MaoSYC21} formulated this task as a multi-turn machine reading comprehension task and solve it with machine reading comprehension frameworks. Recently, \cite{DBLP:conf/acl/XuCB20} had propose a span-level model to extract ATs and OTs first and then predict the sentiment relation for each (AT, OT) pairs.

\begin{figure*} [t]
	\centering
	\includegraphics[scale=0.6]{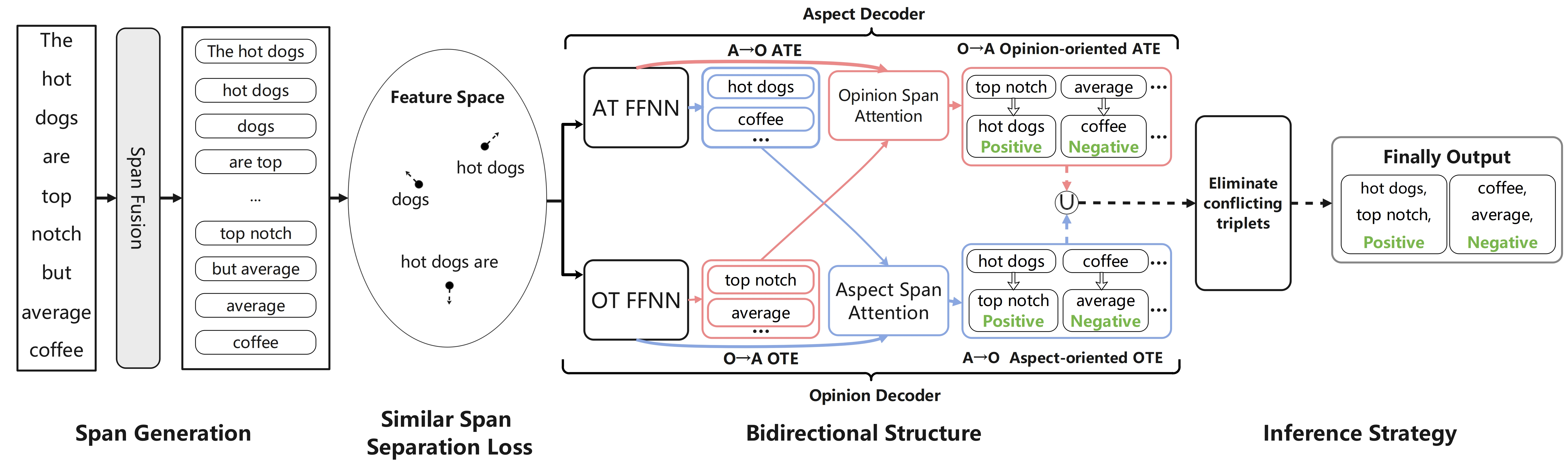}
	\caption{The overall architecture of our span-level bidirectional network. The blue arrows and modules as well as red arrows and modules indicate the extraction of aspect-to-opinion direction and the opinion-to-aspect direction, respectively. The process shown in the dotted line only proceeds in the inference.}
	\label{framework}
\end{figure*}

\section{Methodology}
As shown in Figure \ref{framework}, our network consists of four parts: span generation, similar span separation loss, bidirectional structure, and the inference strategy. In the following subsections, we first give the definition of ASTE tasks and then detail our network structure.

\subsection{Task Definition} 
For a sentence $S=\left\{w_{1},w_{2},\dots,w_{n}\right\}$ consisting $n$ words, the goal of the ASTE task is to extract a set of aspect sentiment triplets $\mathcal{T}=\left\{{(a,o,c)}_{k}\right\}_{k=1}^{|\mathcal{T}|}$ from the given sentence $S$, where $(a,o,c)$ refers to (aspect term, opinion term, sentiment polarity) and $c\in\left\{Positive,Neutral,Negative\right\}$.

\subsection{Span Generation} \label{span generation}
Given a sentence $S$ with $n$ tokens, there are $m$ possible spans in total. Each span $\mathbf{s}_{i}=\left\{w_{start(i)},\dots,w_{end(i)}\right\}$ is defined by all the tokens from $start(i)$ to $end(i)$ inclusive, and the maximum length of span $\mathbf{s}_{i}$ is $l_{s}$:
\begin{equation}
	1 \leq start(i) \leq end(i) \leq n
\end{equation}
\begin{equation}
	end(i) - start(i) \leq l_{s}
\end{equation}

To obtain span representations, we need to get the token-level representations first. In this paper, we utilize BERT \cite{devlin2018bert} as a sentence encoder to obtain token-level contextualized representations $\left\{\mathbf{h}_{1},\mathbf{h}_{2},\dots,\mathbf{h}_{n}\right\}$ of the given sentence $S$. Then, the token-level representations are combined by max pooling. Note that various methods can be applied to generate the representations for spans, the effectiveness of these span generation methods will be investigated in the ablation study in Appendix. We define the representation of span $\mathbf{s}_{i}$ as:
\begin{equation}
	\mathbf{g}_{i}=Max\left(\mathbf{h}_{start}(i), \mathbf{h}_{start+1}(i),\dots,\mathbf{h}_{end}(i)\right)
\end{equation}
where $Max$ represents max pooling.

\subsection{Similar Span Separation Loss}
After generating the representation of span, most previous models directly use the span representations for downstream tasks. However, enumerating all possible spans in a sentence inevitably generates lots of spans that have same tokens with each other, and the model may suffer from the limitations in processing these similar spans due to their adjacent distribution. To separate the spans with similar distributions, we propose a similar span separation loss based on KL divergence to separate similar spans, as shown in Figure \ref{framework}. The similar span separation loss is defined as:
\begin{equation}
	\begin{split}
		KL(\mathbf{g}_{i}||G_{i}) &= \\ \sum_{j}^{G_{i}}&softmax(\mathbf{g}_{i}) log \frac{softmax(\mathbf{g}_{i})}{softmax(\mathbf{g}_{j})}
	\end{split}
\end{equation}
\begin{equation}
	\begin{split}
		KL(G_{i}||\mathbf{g}_{i}) &= \\ \sum_{j}^{G_{i}} &softmax(\mathbf{g}_{j}) log \frac{softmax(\mathbf{g}_{j})}{softmax(\mathbf{g}_{i})}
	\end{split}
\end{equation}
\begin{equation}
	\begin{split}
		\mathcal{J}_{KL} &= \\ \sum_{i}^{m} &log(1 + \frac{2}{KL(G_{i}||\mathbf{g}_{i}) + KL(\mathbf{g}_{i}||G_{i})})
	\end{split}
\end{equation}
where $G_{i}$ indicates the set of the representations of spans which share at least one token with $\mathbf{s}_{i}$. Note that we have not directly used the KL divergence as the separation loss but in combination with the $log(1+1/x)$ function to achieve the effect that when KL divergence is small the separation loss is large and vice versa.

\subsection{Bidirectional Structure} \label{BCS}
As the aspect sentiment triplet can be triggered by an aspect terms or an opinion terms, we propose a bidirectional structure to decode the span representations. As shown in Figure \ref{framework}, the bidirectional structure consists of an aspect decoder and an opinion decoder. The details of each component in the bidirectional structure are given in the following subsections.

\subsubsection{Aspect-to-opinion Direction}
In aspect-to-opinion direction (Blue arrows and modules in Figure \ref{framework}), the aspect decoder aims to extract all ATs along with their sentiment from the sentence. We can obtain the confidence score as well as the probability of the sentiment of AT as follows:
\begin{equation} \label{aspect FFN}
	u^{a}_{i}=FFNN_{a}\left(\mathbf{g}_{i}, \theta_{a}\right)
\end{equation}
\begin{equation}
	q^{a\rightarrow o,a}_{i}=\mathbf{w}_{a\rightarrow o,a}u^{a}_{i}
\end{equation}
\begin{equation}
	\mathbf{p}^{a\rightarrow o,a}_{i}=softmax(q^{a\rightarrow o,a}_{i})
\end{equation}
where $FFNN_{A}$ represents the FFNN of aspect decoder, $\theta_{a}$ is the parameter for the FFNN, $\mathbf{w}_{a\rightarrow o,a}\in\mathbb{R}^{m\times c^{0}}$ is a trainable weight vector, and $c^{0}\in\left\{Valid, Invalid\right\}$ is the number of categories. 

Then, giving a set $G_{a}$ of original span representations of all valid ATs $\mathbf{g}^{a}_{j}\in G_{a}$, we apply the opinion decoder to identify all OTs along with their sentiment for each particular valid AT by exploiting attention mechanism. Similarly, we obtain the probability distribution of the OT's sentiment along with its confidence score via:
\begin{equation} \label{opinion FFN}
	u^{o}_{i}=FFNN_{o}\left(\mathbf{g}_{i}, \theta_{o}\right)
\end{equation}
\begin{equation} 			
	\alpha^{a\rightarrow o}_{i,j}=\frac{\exp(u^{o}_{i})} {\exp(\mathbf{g}^{a}_{j})}
\end{equation}
\begin{equation}
	q^{a\rightarrow o, o}_{i,j}=\mathbf{w}_{a\rightarrow o,o}\left(u^{o}_{i} + \alpha^{a\rightarrow o}_{i,j} \cdot \mathbf{g}^{a}_{j} \right)
\end{equation}
\begin{equation}
	\mathbf{p}^{a\rightarrow o, o}_{i,j}=softmax(q^{a\rightarrow o, o}_{i,j})
\end{equation}
where $FFNN_{o}$ represents the FFNN of opinion decoder, $\theta_{o}$ is the parameter for the FFNN, $\mathbf{w}_{a\rightarrow o,o}\in\mathbb{R}^{m\times c^{*}}$ is a trainable weight vector, and $c^{*}\in\left\{Positive,Neutral,Negative,Invalid\right\}$ is the number of sentiment polarity. Furthermore, we define the loss of aspect-to-opinion direction as:
\begin{equation}
	\begin{split}
		\mathcal{J}_{a\rightarrow o}&=-\sum_{i} y^{a\rightarrow o,a}_{i} \log\left(q^{a\rightarrow o,a}_{i}\right) \\ &- \sum_{i} \sum_{j}^{G_{a}} y^{a\rightarrow o,o}_{i,j} \log\left(q^{a\rightarrow o,o}_{i, j}\right)
	\end{split}
\end{equation}
where $y^{a\rightarrow o,a}_{i}$ and $y^{a\rightarrow o,o}_{i,j}$ are ground truth labels of the sentiments for AT and OT given a specific valid AT, respectively.

\subsubsection{Opinion-to-aspect Direction}
As for opinion-to-aspect direction (Red arrows and modules in Figure \ref{framework}), the opinion decoder is deployed first to extracts all the OTs along with their sentiment from the sentence. To minimize the number of model parameters, the opinion decoder in both aspect-to-opinion and opinion-to-aspect directions shares the FFNN features, as described in Equation \eqref{opinion FFN}. The probability distribution of the sentiments of OTs as well as the confidence scores can be obtained as:
\begin{equation}
	q^{o\rightarrow a,o}_{i}=\mathbf{w}_{o\rightarrow a,o}u^{o}_{i}
\end{equation}
\begin{equation}
	\mathbf{p}^{o\rightarrow a,o}_{i}=softmax(q^{o\rightarrow a,o}_{i})
\end{equation}
where $\mathbf{w}_{o\rightarrow a,o}\in\mathbb{R}^{m\times c^{0}}$ is a trainable weight vector. 

Given a set $G_{o}$ if original span representations of all valid OTs $\mathbf{g}^{o}_{j}\in G_{o}$, the aspect decoder is deployed to identify the ATs and their sentiment for each particular valid OTs. Note that the aspect decoder in opinion-to-aspect direction also shares same FFNN features described in Equation \eqref{aspect FFN} with the aspect decoder in aspect-to-opinion direction. The logits of ATs and their confidence scores in opinion-to-aspect direction can be obtained by:
\begin{equation} 			
	\alpha^{o\rightarrow a}_{i,j}=\frac{\exp(u^{a}_{i})}  {\exp(\mathbf{g}^{o}_{j})}
\end{equation}
\begin{equation}
	q^{o\rightarrow a,a}_{i,j}=\mathbf{w}_{o\rightarrow a,a}\left(u^{a}_{i} + \alpha^{o\rightarrow a}_{i,j} \cdot \mathbf{g}^{o}_{j} \right)
\end{equation}
\begin{equation}
	\mathbf{p}^{o\rightarrow a,a}_{i,j}=softmax(q^{o\rightarrow a,a}_{i,j})
\end{equation}
where $\mathbf{w}_{o\rightarrow a,a}\in \mathbb{R}^{m\times c^{*}}$ is a trainable weight vector. 

Finally, the loss for opinion-to-aspect direction is defined as:
\begin{equation}
	\begin{split}
		\mathcal{J}_{o\rightarrow a}&=-\sum_{i} y^{o\rightarrow a,o}_{i} \log\left(q^{o\rightarrow a,o}_{i}\right) \\ 
		&- \sum_{i} \sum_{j}^{G_{o}} y^{o\rightarrow a,a}_{i,j} \log\left(q^{o\rightarrow a,a}_{i,j}\right)
	\end{split}
\end{equation}
where $y^{o\rightarrow a,o}_{i}$ and $y^{o\rightarrow a,a}_{i,j}$ are the ground truth labels. Then, we combine the above loss functions to form the loss objective of the entire model:
\begin{equation}
	\mathcal{J}= \mathcal{J}_{KL} +  \mathcal{J}_{a\rightarrow o} + \mathcal{J}_{o\rightarrow a}
\end{equation}

\begin{algorithm}[t] 
	\caption{Inference Strategy}
	\renewcommand{\algorithmicrequire}{\textbf{Input:}}
	\renewcommand{\algorithmicensure}{\textbf{Output:}}
	\begin{algorithmic}[1] \label{algorithm}
		\REQUIRE $\mathcal{T}_{a\rightarrow o} $,$ \mathcal{T}_{o\rightarrow a}$ \\
		$\mathcal{T}_{a\rightarrow o}$ denotes the triplet extraction results in aspect-to-opinion direction \\
		$\mathcal{T}_{o\rightarrow a}$ denotes the triplet extraction results in opinion-to-aspect direction
		\STATE Get the overall triplets in both extract directions $\mathcal{T} = \mathcal{T}_{a\rightarrow o} \cup \mathcal{T}_{o\rightarrow a}$
		\FOR {$ t_{i} \in \mathcal{T}$}
		\FOR {$t_{j}\in \left(\mathcal{T} - \left\{ t_{i} \right\} \right)$}
		\STATE $t_{i}= (a_{i},o_{i},c_{i},s_{i}), t_{j}= ( a_{j}, o_{j},c_{j},s_{j})$, $s_{i}$ and $s_{j}$ are the confidence score of the corresponding triplets
		\IF {$a_{i} \cap a_{j} \neq \varnothing$ \AND $o_{i} \cap o_{j} \neq \varnothing$}
		\IF {$s_{i} > s_{j}$}
		\STATE $\mathcal{T} =  \mathcal{T} - \left\{ t_{j} \right\}$
		\ELSE
		\STATE $\mathcal{T} =  \mathcal{T} - \left\{ t_{i} \right\}$
		\ENDIF
		\ENDIF
		\ENDFOR
		\ENDFOR
		\RETURN $\mathcal{T}$
	\end{algorithmic}
\end{algorithm}

\subsection{Inference} \label{infer}
In contrast to the mutual exclusivity of the triplets in the token-level method, span-level model cannot guarantee that there are no conflicts between any two triples. Therefore, we propose an inference strategy to eliminate the potential conflicting triplets during the inference process. As illustrated in Algorithm \ref{algorithm}, We first combine the extraction results in both directions by taking the union set $\mathcal{T}$ (line 1). Afterwards, for each pair of triplets in the overall triplets set $\mathcal{T}$ that have duplicates in both aspect $a$ and opinion $o$ (line 5), the conflicting results are eliminated by discarding the triplets with lower confidence scores $s$ (line 6-9). Note that in the condition of determining whether two triplets conflict with each other (line 5), the determination of whether the union set is empty is performed on the position index, rather than on the tokens.

\section{Experiments}

\subsection{Datasets}
\begin{table}[t]
	\scriptsize
	\centering
	\begin{tabular}{cc|cccccc}
		\hline
		\multicolumn{2}{c|}{\multirow{2}{*}{Datasets}}      & \multirow{2}{*}{\#S} & \multirow{2}{*}{POS} & \multirow{2}{*}{NEU} & \multirow{2}{*}{NEG} & \multirow{2}{*}{\#SW} & \multirow{2}{*}{\#MW} \\
		\multicolumn{2}{c|}{}                               &                           &                           &                          &                           &                              &                            \\ \hline
		\multicolumn{1}{c|}{\multirow{3}{*}{14LAP}} & Train & 1266                      & 1692                      & 166                      & 480                       & 1586                         & 752                        \\
		\multicolumn{1}{c|}{}                       & Dev   & 310                       & 404                       & 54                       & 119                       & 388                          & 189                        \\
		\multicolumn{1}{c|}{}                       & Test  & 492                       & 773                       & 66                       & 155                       & 657                          & 337                        \\ \hline
		\multicolumn{1}{c|}{\multirow{3}{*}{14RES}} & Train & 906                       & 817                       & 126                      & 517                       & 824                          & 636                        \\
		\multicolumn{1}{c|}{}                       & Dev   & 219                       & 169                       & 36                       & 141                       & 190                          & 156                        \\
		\multicolumn{1}{c|}{}                       & Test  & 328                       & 364                       & 63                       & 116                       & 291                          & 252                        \\ \hline
		\multicolumn{1}{c|}{\multirow{3}{*}{15RES}} & Train & 605                       & 783                       & 25                       & 205                       & 678                          & 335                        \\
		\multicolumn{1}{c|}{}                       & Dev   & 148                       & 185                       & 11                       & 53                        & 165                          & 84                         \\
		\multicolumn{1}{c|}{}                       & Test  & 322                       & 317                       & 25                       & 143                       & 297                          & 188                        \\ \hline
		\multicolumn{1}{c|}{\multirow{3}{*}{16RES}} & Train & 857                       & 1015                      & 50                       & 329                       & 918                          & 476                        \\
		\multicolumn{1}{c|}{}                       & Dev   & 210                       & 252                       & 11                       & 76                        & 216                          & 123                        \\
		\multicolumn{1}{c|}{}                       & Test  & 326                       & 407                       & 29                       & 78                        & 344                          & 170                        \\ \hline
	\end{tabular}
	\caption{Statistics of the datasets. `\#S' denotes the numbers of sentence, `POS',`NEU', and `NEG' denote the numbers of positive, neutral, and negative triplets respectively. `\#SW' denotes the numbers of triplets where the ATs and OTs are single word spans. `\#MW' denotes the numbers of triplets that at least one of the ATs or the OTs are multi-word spans.}
	\label{datasets}
\end{table}	

To verify the effectiveness of our network, we conduct experiments on four benchmark datasets\footnotemark \cite{DBLP:conf/emnlp/XuLLB20} , which are constructed based on the original SemEval ABSA Challenges and the datasets of \cite{DBLP:conf/naacl/FanWDHC19}. Table \ref{datasets} lists the statistics of these datasets.

$\footnotetext{https://github.com/xuuuluuu/SemEval-Triplet-data/tree/master/ASTE-Data-V2-EMNLP2020}$
\subsection{Experimental Setting}
We adopt the cased base version of \textbf{BERT} \cite{devlin2018bert} in our experiments, which contains 110M parameters. During training, we use AdamW \cite{DBLP:journals/corr/abs-1711-05101} to optimize the model parameters. The fine-tuning rate for BERT and the learning rate for other models are set to 1e-5 and 1e-4, respectively. Meanwhile, the mini-batch size is set to 16 and the dropout rate is set to 0.1. The maximum length of generated spans is set to 8. We train our framework in a total of 120 epochs on a NVIDIA Tesla V100 GPU. 
\subsection{Evaluation}
To comprehensively evaluate the performance of different methods, we use \emph{precision}, \emph{recall}, \emph{F1-score} as the evaluation metrics. The extracted ATs and OTs are considered correct if and only if predicted spans exactly match the ground truth spans. In the experiments, we select the testing results when the model achieves the best performance on the development set.
\begin{table*}[h]
	\footnotesize
	\centering
	\resizebox{\textwidth}{!}{\begin{tabular}{c|ccc|ccc|ccc|ccc}
			\hline
			& \multicolumn{3}{c|}{14LAP}                       & \multicolumn{3}{c|}{14RES}               & \multicolumn{3}{c|}{15RES}                       & \multicolumn{3}{c}{16RES}                        \\ \cline{2-13} 
			& P              & R              & F1             & P      & R              & F1             & P              & R              & F1             & P              & R              & F1             \\ \hline
			PENG-two-stage & 40.40          & 47.24          & 43.50          & 44.18  & 62.99          & 51.89          & 40.97          & 54.68          & 46.79          & 46.76          & 62.97          & 53.62          \\
			JETt           & 51.48          & 42.65          & 46.65          & 70.20  & 53.02          & 60.41          & 62.14          & 47.25          & 53.68          & \textbf{72.12} & 57.20          & 63.41          \\
			JETo           & 58.47          & 43.67          & 50.00          & 67.97  & 60.32          & 63.92          & 58.35          & 51.43          & 54.67          & 64.77          & 61.29          & 62.98          \\
			GTS-BERT       & 57.52          & 51.92          & 54.58          & 70.92  & 69.49          & 70.20          & 59.29          & 58.07          & 58.67          & 68.58          & 66.60          & 67.58          \\
			Dual-MRC       & 57.39          & 53.88          & 55.58          & 71.55  & 69.14          & 70.32          & 63.78          & 51.87          & 57.21          & 68.60          & 66.24          & 67.40          \\
			B-MRC          & \textbf{70.89*}         & 50.20*         & 58.78*         & 75.41* & 64.04*         & 69.26*         & 69.83*         & 56.04*         & 58.74*         & 69.03*         & 66.02*         & 67.49*         \\
			Span-ASTE      & 63.44          & 55.84          & 59.38          & 72.89  & 70.89          & 71.85          & 62.18          & \textbf{64.45} & 63.27          & 69.45          & 71.17          & 70.26          \\
			Ours           & 65.68 & \textbf{59.88} & \textbf{62.65} & \textbf{76.36}  & \textbf{72.43} & \textbf{74.34} & \textbf{69.93} & 60.41          & \textbf{64.82} & 71.59          & \textbf{72.57} & \textbf{72.08} \\ \hline
	\end{tabular}}
	\caption{Precision (\%), Recall (\%) and F1 score (\%) on the test set of the ASTE tasks. State-of-the-art results are marked bold. * indicates that the result is reproduced by us.}
	\label{main result}
\end{table*}

\subsection{Baselines}
To demonstrate the effectiveness of our network, we compare our method with the following baselines:
\begin{itemize}
	
	\item \textbf{Peng-two-stage} \cite{DBLP:conf/aaai/PengXBHLS20} is a two-stage pipeline model. Peng-two-stage extracts both aspect-sentiment pairs and opinion terms in the first stage. In the second stage, Peng-two-stage pairs up the extraction results into triplets via an relation classifier.
	
	\item \textbf{JET} \cite{DBLP:conf/emnlp/XuLLB20} is an end-to-end model which proposes a novel position-aware tagging scheme to jointly extracting the triplets. It also designs factorized feature representations so as to effectively capture the interaction among the triplet factors. 
	
	\item \textbf{GTS} \cite{DBLP:journals/corr/abs-2010-04640} is an end-to-end model which formulates ASTE as a unified grid tagging task. It first extracts the sentiment feature of each token, and then
	gets the initial prediction probabilities of toke pairs based on these token-level features. It also designs an inference strategy to exploit the potential mutual indications between different opinion factors and performs the final prediction.
	
	\item \textbf{Dual-MRC} \cite{DBLP:conf/aaai/MaoSYC21} is a joint training model which consists of two machine reading comprehensions. One of the MRC is for aspect term extraction, and another is for aspect-oriented opinion term extraction and sentiment classification.
	
	\item \textbf{B-MRC} \cite{chen2021bidirectional} formalizes the ASTE task as a multi-turn machine reading comprehension task, and proposes three types of queries to extract targets, opinions and the sentiment polarities of aspect-opinion pairs, respectively.
	
	\item \textbf{Span-ASTE} \cite{DBLP:conf/acl/XuCB20} considers all possible spans in a sentence and build the interaction between the whole spans of aspect terms and opinion terms when predicting their sentiment relation. They also propose a dual-channel span pruning strategy to ease the high computational cost caused by span enumeration.
	
\end{itemize}

\begin{table}[t]
	\footnotesize
	\centering
	\resizebox{\linewidth}{!}{
		\begin{tabular}{c|ccccc}
		\hline
		& KL Loss  & JS Loss  &EM Loss & CS Loss & no Loss \\ \hline
		Ours      & \textbf{62.65} & 62.34          & 61.17         & 61.51         & 60.82         \\
		w/o IS  & \textbf{61.48} & 61.05          & 60.40         & 60.82         & 60.08         \\ \hline
		Ours$_{a\rightarrow o}$     & \textbf{62.14} & 61.86          & 60.08         & 61.70         & 60.66         \\
		w/o IS & 61.09          & \textbf{61.18} & 60.08         & 60.67         & 58.78         \\ \hline
		Ours$_{o\rightarrow a}$     & \textbf{62.38} & 62.03          & 60.88         & 61.70         & 60.68         \\
		w/o IS & \textbf{61.75} & 60.15          & 59.90         & 60.41         & 59.38         \\ \hline
		\end{tabular}
	}
	\caption{Experimental results of the ablation study in 14LAP dataset (\emph{F1-score}, \%). `w/o IS' denotes not using the inference strategy described in Section \ref{infer}. `KL Loss', `JS Loss', `EM Loss', and `CS Loss' denote that similar span separation loss is constructed base on KL divergence, JS divergence, Euclidean Metric, and Cosine Similarity, respectively. `no Loss' means no application of similar span separation loss.}
	\label{directions} 
\end{table}

\subsection{Main Results}

Table \ref{main result} reports the results of our framework and baseline models. According to the results, our framework achieves state-of-the-art performance on all datasets. Specifically, our framework surpasses the best baselines by an average of 2.3 \emph{F1-score} on ASTE. This result demonstrates that our framework can take advantage of bidirectional decoding and efficiently distinguish the span representation. Although some of the recall scores are slightly lower than Span-ASTE, the increase in precision significantly outperforms the previous baselines in most datasets, which shows the higher prediction accuracy of our network. It is worth noting that BMRC and Dual-MRC achieve better performance than JET and PENG-two-stage. This is probably because BMRC and Dual-MRC formalize the ASTE task as a multi-turn machine reading comprehension task and benefit from asking the model questions. Unlike those approaches, Span-ASTE and our method both utilize the span-level interactions to handle the ASTE task and avoid the cascading errors. Moreover, our model outperforms Span-ASTE because our method identify the triplets from both aspect-to-opinion and opinion-to-aspect directions, rather than matching each aspect span and opinion span.  Besides, our network also take advantage of similar span separation loss and inference strategy to overcome the drawback of mutual exclusivity absence among spans.
\subsection{Ablation Study} \label{Ablation Study} 

To validate the origination of the significant improvement in our network, we conduct ablation experiments on 14LAP datasets. As shown in Table \ref{directions}, our bidirectional model yields better results than unidirectional models, which clearly indicates the superiority of the collaboration in both two directions on decoding the span representations. And the inference results from opinion terms to aspect terms are better than the other direction, which may due to the simplicity of extracting opinion terms in the 14LAP dataset. 

Moreover, the inference strategy has exhibited the enhancement on model performance. However, the improvement brought by the inference strategy is not significant, because conflicting triplets tend to exist among multi-token results, and only a small percentage of triplets containing multi-token terms in 14LAP dataset. We believe the effect of the inference strategy will be more obvious in datasets enriched with multi-token triplets.

In addition, to demonstrate the effectiveness of our proposed similar span separation loss based on KL divergence, we further design similar range separation losses based on JS divergence, Euclidean distance and cosine similarity. The experimental results show that all these loss functions have a boosting effect on our network, and the separation loss based on KL divergence performs the best. Note that numerous similarity measures can be used to separate similar spans, among which there may be some better measures that can bring more improvement to the model.
\subsection{Effect of Entity Length}
\begin{figure}[t]
	\centering
	\subfloat[ATE on 14LAP]{
		\includegraphics[scale=0.712]{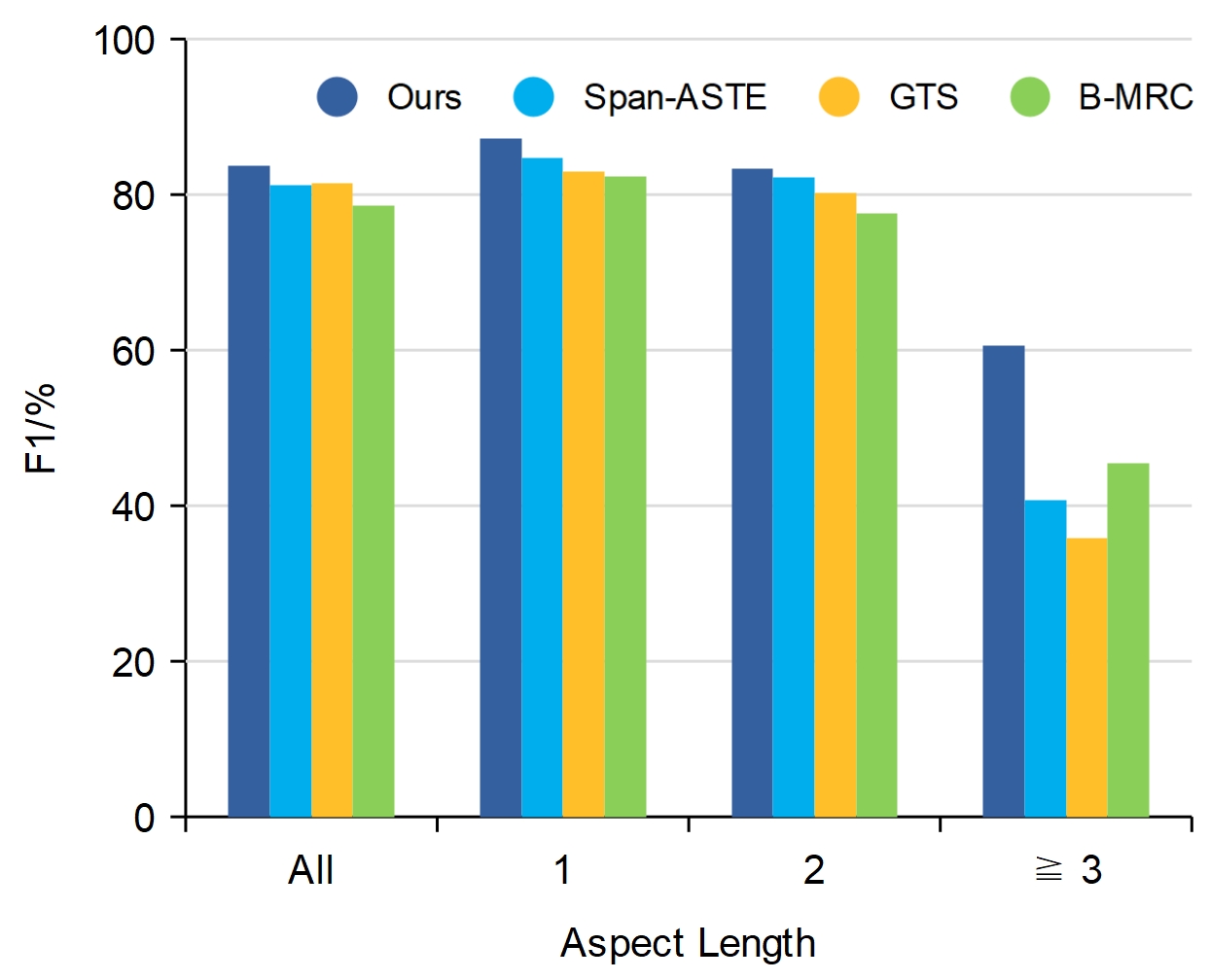}}
	\hfil
	\subfloat[OTE on 14LAP]{
		\includegraphics[scale=0.712]{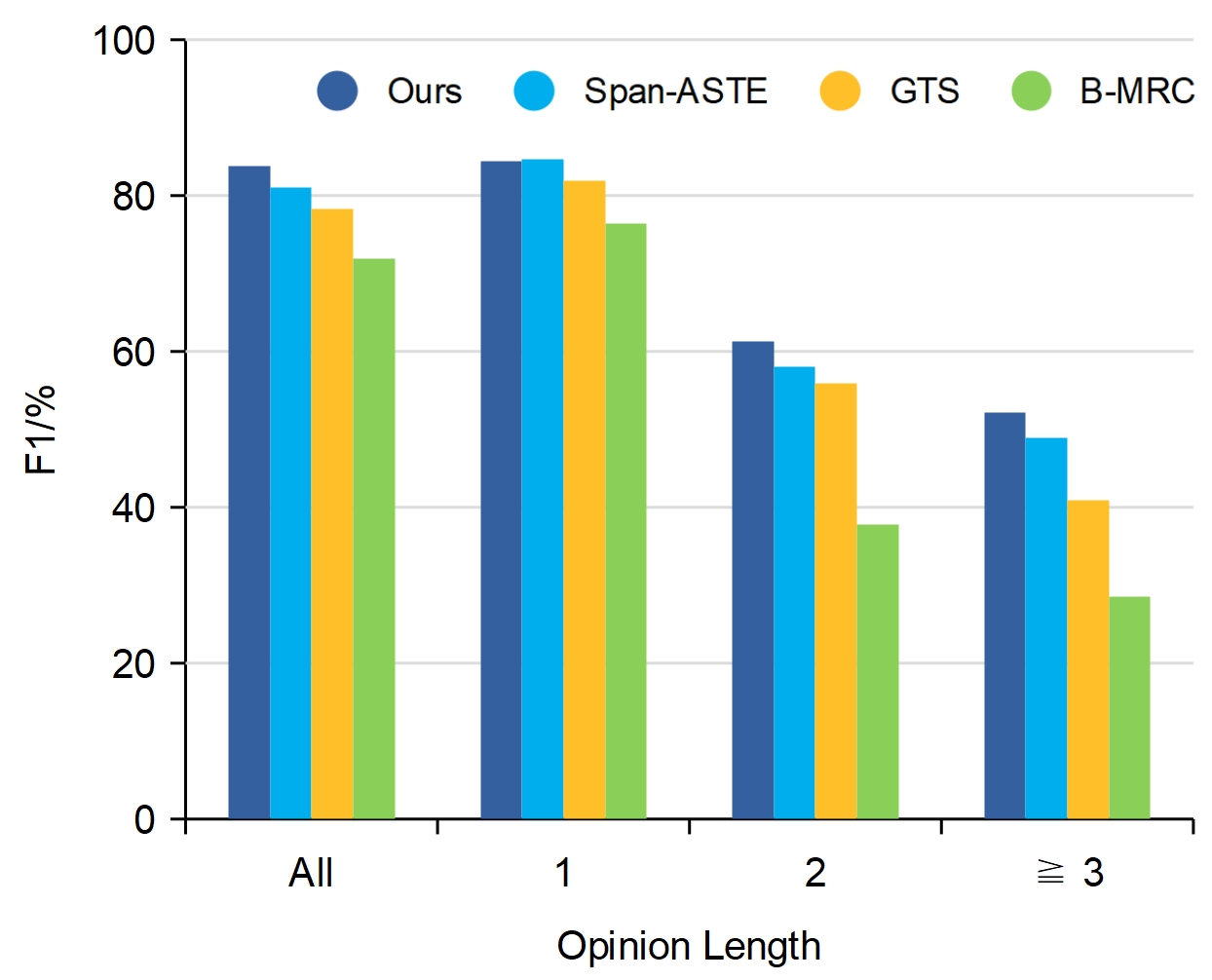}}
	\hfil
	\caption{Effects of entity length for aspect term extraction and opinion term extraction (\emph{F1-score}, \%).}
	\label{Target length}
\end{figure}

To investigate the performance of different methods on the ATE and OTE with different entity lengths, we report the F1 scores of our framework, Span-ASTE, GTS, and B-MRC on the extraction task with different lengths of entities. The results are illustrated in Figure \ref{Target length}. As the entity length increases, the performance gap between our framework and other models becomes more obvious. Since our method directly models span-level feature for each entity and and alleviates the drawback of no mutual exclusivity among spans, our method will not be greatly affected with entity lengths increasing. In fact, most of the contribution to the improvement in our model comes from the performance in multi-token entities.

\subsection{Effect of Multiple Triplets}

\begin{table}[htpb]
	\footnotesize
	\centering
	\resizebox{\linewidth}{!}{
		\begin{tabular}{c|cccccc}
			\hline
			& All            & 1              & 2              & 3              & 4              & $\ge$ 5            \\ \hline
			Ours       & \textbf{62.65} & \textbf{62.80}  & \textbf{63.54} & \textbf{64.81} & \textbf{57.14} & \textbf{60.46} \\
			Span-ASTE & 59.65          & 61.18 			& 61.01          & 61.67          & 56.00          & 35.00          \\
			GTS       & 58.57          & 58.21          & 60.26          & 65.00          & 56.57          & 30.77          \\
			B-MRC     & 58.78          & 58.10          & 64.75          & 55.31          & 35.56          & 0.00           \\ \hline
		\end{tabular}
	}
	\caption{Effects of multiple triplets in a sentence in 14LAP (\emph{F1-score}, \%).}
	\label{Triplet number}
\end{table}

To further verify the ability of our framework to handle multiple triplets, we compare the performance of our network and other baselines on ASTE task with different number of triplets in the sentences, and the results are shown in Table \ref{Triplet number}. We divide the sentences in 14LAP testset into 5 subclasses. Each subclass contains sentences with 1, 2, 3, 4, or $\geq$ 5 triplets, respectively. When extracting triplets from sentences that contain 1 or 2 triplets, the performance of our framework is competitive to other models. However, when the number of triplets increases, the performance of Span-ASTE, GTS, and B-MRC decrease significantly, while the performance of our network remains stable or even slightly increases. These experimental results demonstrate the efficiency and stability of our framework in handling multiple triplets in a sentence.

\section{Conclusions}
In this work, we propose a span-level bidirectional network for ASTE tasks. This span-level model can take advantages from both aspect-to-opinion and opinion-to-aspect directions. The bidirectional decoding can ensure that either an AT or an OT can trigger an aspect sentiment triplet, which is more in line with human perception. For the shortcoming that mutual exclusivity cannot be guaranteed among spans, we deploy the similar span separation loss to guide the model in discriminating similar spans. We further design an inference strategy to eliminate conflicting triplet results that are specific to span-level models. The experimental results demonstrate that our network significantly outperforms the compared baselines and achieves state-of-the-art performance. 

\section*{Limitations}
Although in the previous section we showed the advanced performance of the network we designed, there are still some weaknesses in our model.

\begin{table}[htpb]
	\footnotesize
	\centering
	\begin{tabular}{c|cc}
		\hline
		& MACs(G) & Params(M)  \\ \hline
		Ours      & 120.044 & 129.884   \\
		Span-ASTE & 444.55  & 110       \\
		B-MRC     & 19.624  & 85.611    \\
		GTS       & 520.765 & 88.006    \\ \hline
	\end{tabular}
\caption{Efficiency Comparison.}
\label{Efficiency}
\end{table}

First, our model uses spans as input, and enumerating all possible spans inevitably increases the input size of the model, so span-level models tend to have larger computations than token-level models. As shown in Table \ref{Efficiency}, our network requires about 6 times more floating-point computations than the B-MRC model. While the Span-ASTE and GTS models require more computation, this is because Span-ASTE needs to match every aspect terms and opinion terms and GTS model extracts triplets by classifying the internal elements of a square matrix consisting of sentences in rows and columns.

Second, to reduce the input size of the model, we set the maximum span length of the spans to 8 to include as many potential aspect terms and opinion terms as possible. However, in some datasets with long extraction targets, the span-level model must increase the maximum span limit, thus affecting the performance of the model. Therefore, our model is suitable only for the tasks of extracting short targets.

Third, both the similar span separation loss and inference strategy proposed in this paper are used to alleviate the shortcoming  of the missing mutual exclusivity in span-level models, while the inputs of token-level models are naturally mutually exclusive. So the similar range separation loss and inference strategies are not applicable to token-level models.

\section*{Ethics Statement}
	This article does not contain any study with human participants or animals performed by any of the authors. And all authors declare that they have no known competing financial interests or personal relationships that could have appeared to influence the work reported in this paper.

\bibliography{anthology}
\bibliographystyle{acl_natbib}

\appendix

\end{document}